\documentclass[letterpaper, 10 pt, conference]{ieeeconf}  

\IEEEoverridecommandlockouts                              

\title{\LARGE \bf
Self-supervised Cloth Reconstruction \\via Action-conditioned Cloth Tracking
}

\author{Zixuan Huang$^{1}$, Xingyu Lin$^{1}$, David Held$^{1}$
\thanks{$^{1}$ All authors are affiliated with Robotics Institute, Carnegie Mellon University, 5000 Forbes Ave, USA.
        {\tt\small \{zixuanhu,xlin3,dheld\}@andrew.cmu.edu}}%
}

\usepackage{times}

\usepackage{multicol}
\usepackage{graphicx}
\usepackage{hyperref}
\usepackage{amsmath}
\usepackage{cleveref}
\usepackage[ruled,vlined,linesnumbered]{algorithm2e}
\usepackage[font=footnotesize,labelfont=bf]{caption}
\usepackage{subcaption}
\usepackage{bbm}
\usepackage{float}
\usepackage{color, soul} 
\usepackage{wrapfig,lipsum,booktabs}

\usepackage{array}
\usepackage[utf8]{inputenc}
\usepackage{amsmath,  amssymb}
\usepackage{booktabs}
\usepackage{dsfont}
\usepackage{diagbox}
\usepackage{multirow}

\DeclareMathOperator*{\argmin}{arg\,min}

\usepackage{color, soul}
\begin{document}

\maketitle
\thispagestyle{empty}
\pagestyle{empty}


\begin{abstract}
State estimation is one of the greatest challenges for cloth manipulation due to cloth's high dimensionality and self-occlusion. Prior works propose to identify the full state of crumpled clothes by training a mesh reconstruction model in simulation. However, such models are prone to suffer from a sim-to-real gap due to differences between cloth simulation and the real world. In this work, we propose a self-supervised method to finetune a mesh reconstruction model in the real world. Since the full mesh of crumpled cloth is difficult to obtain in the real world, we design a special data collection scheme and an action-conditioned model-based cloth tracking method to generate pseudo-labels for self-supervised learning. By finetuning the pretrained mesh reconstruction model on this pseudo-labeled dataset, we show that we can improve the quality of the reconstructed mesh without requiring human annotations, and improve the performance of downstream manipulation task. 
More visualizations and results can be found on our \href{https://sites.google.com/view/ss-mesh-recon/home}{project website}.
\end{abstract}

\section{Introduction}

Despite the ubiquitous presence of cloth in the real-world, manipulating it with a robot remains a difficult task. Specifically, the high dimensionality and self-occlusion of cloth pose significant challenges for precise state estimation. Prior works~\cite{jiang2020bcnet,danvevrek2017deepgarment,saito2021scanimate,su2022deepcloth, chi2021garmentnets} 
try to reconstruct the full mesh of cloth from RGB or depth observations; the mesh reconstruction model can be used for robot cloth manipulation~\cite{li2014real, li2014recognition,mariolis2015pose,li2015folding,mariolis2015pose, huang2022mesh}. However, the mesh reconstruction model is typically  trained in simulation and suffers from a sim2real gap between simulated cloth and real cloth.
One approach to mitigate the distribution shift from sim2real is to finetune the model with real world data. On the other hand, obtaining the ground-truth full mesh of crumpled clothes is extremely challenging, because the occluded regions are not observable; this presents a challenge for real-world finetuning.


In this work, we present a self-supervised method that leverages a dynamics model and test-time optimization to generate a pseudo-ground-truth mesh. 
We use a human to collect real-world trajectories via a sequence of pick-and-place actions.
We assume that we are able to  reconstruct the initial cloth mesh (from a flattened configuration); we then track the motion of the cloth during action execution. If we can successfully track the cloth, then we can obtain the full mesh configuration in the real world.
\begin{figure}
    \vspace{-0mm}
    \centering

    \includegraphics[width=\linewidth]{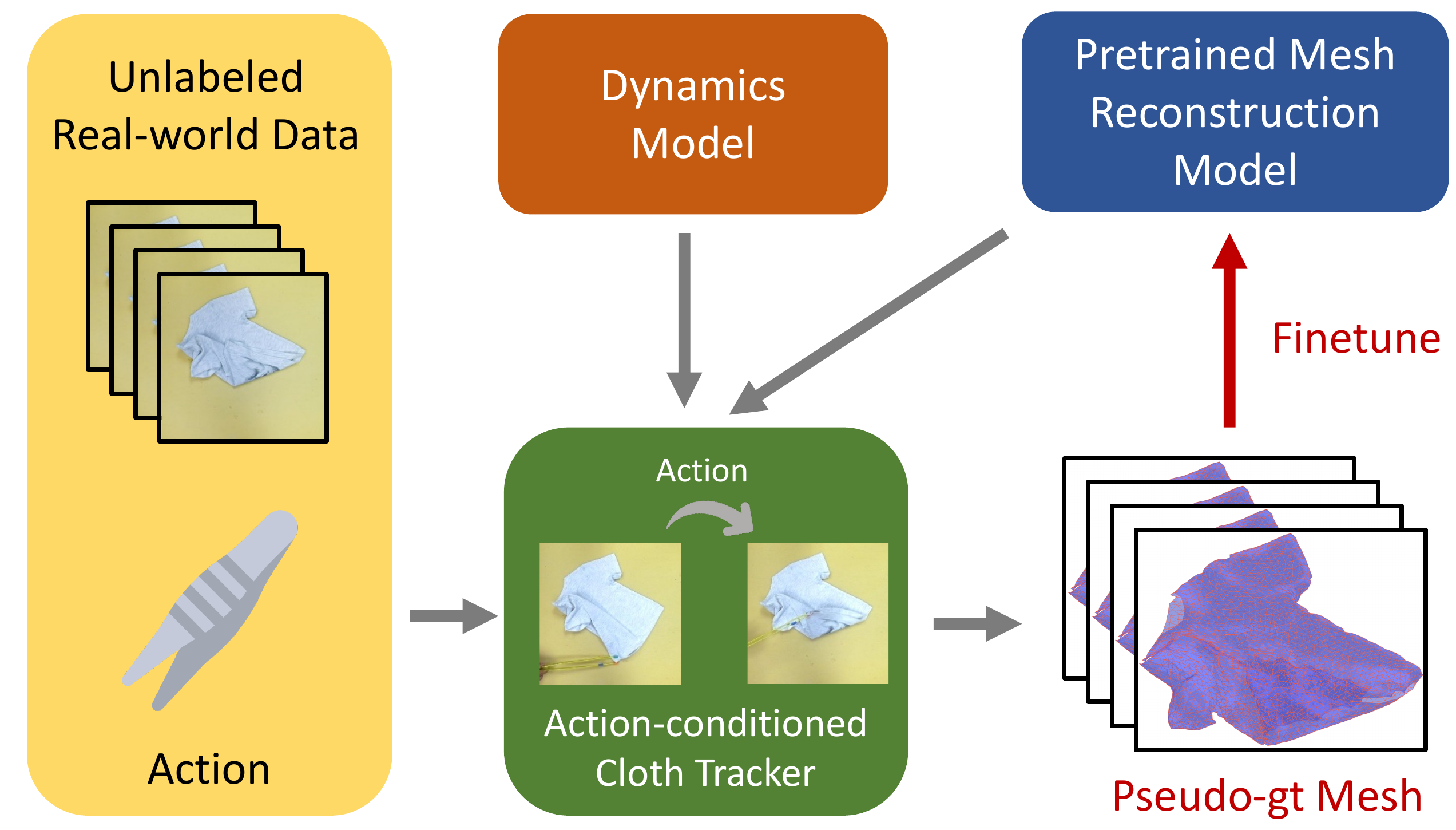}
    \vspace{-3mm}
    \caption{We propose a self-supervised cloth reconstruction method that uses action-conditioned cloth tracking to generate pseudo-labels of the full mesh on real world data. 
    }
    \vspace{-6mm}
    \label{fig:teaser}
\end{figure}

However, tracking the full cloth reliably is a challenging problem. Motivated by the theory of Bayes Filtering, we propose an action-conditioned model-based tracking method. First, we roll out a dynamics (motion) model conditioned on the action to obtain an initial estimate of the motion. This estimate of motion is grounded in physics and accounts for all particles, including the occluded ones.  

However, there will inevitably be gaps between the dynamics model and the real world~\cite{farchy2013humanoid,lund1996simulated,koos2010crossing}, due to incorrect physical parameters and simplified dynamics. To account for dynamics model errors, we further design a test-time optimization method to minimize the discrepancy with the observations at each rollout step, similar to a measurement model of Bayes Filtering.

Our primary contributions are as follows; we: 
\begin{enumerate}
    \item Introduce an action-conditioned cloth tracking method that is robust to occlusions and dynamics errors
    \item Use this tracker for self-supervised fine-tuning in the real world of a mesh reconstruction model. 
    \item Show improved performance of robot cloth flattening.

\end{enumerate}

We use our tracking method to fine-tune a mesh reconstruction model on unlabeled real world data.
Our experiments demonstrate that our method is able to generate plausible pseudo-labels for cloth with complex configurations. 
We also examine the importance of each component of our method using an ablation study.

\section{Related works}
\textbf{Cloth Perception and Manipulation}. Perception and manipulation of clothes has a long history~\cite{Jimenez2017-xn}. Earlier works design heuristic features for specific tasks~\cite{maitin2010cloth,ono1998unfolding,stria2014polygonal}.
More recently, data-driven methods have shown promising results in learning policies~\cite{matas2018sim, wu2019learning, ha2022flingbot} or dynamics model~\cite{seita_fabrics_2020,lin2022learning,huang2022mesh} for cloth smoothing and folding. Particularly for model-based approaches, prior works have shown that learning a dynamics model over the full mesh with occlusion reasoning can significantly improve the planning performance~\cite{huang2022mesh}. While there has been a line of research dedicated to estimating the full mesh of the cloth~\cite{chi2021garmentnets,huang2022mesh,jiang2020bcnet,danvevrek2017deepgarment,patel2020tailornet,saito2021scanimate,su2022deepcloth,li2014real,tanaka2021disruption}, most of these methods are trained on synthetic data and then transfer to the real world, since obtaining mesh data in the real world can be difficult~\cite{bertiche2020cloth3d}. 
As such, this work aims to narrow the sim2real gap by training on real world data collected from cloth tracking. 


\textbf{Deformable Object Tracking}. Numerous deformable object tracking algorithms have been developed, such as template-based tracking~\cite{li2009robust, zollhofer2014real}, simultaneous tracking and reconstruction~\cite{newcombe2015dynamicfusion,slavcheva2018sobolevfusion}, or point set registration~\cite{chui2000new,chui2000feature,myronenko2006non,chi2019occlusion}.
However, these model-free methods are not guaranteed to satisfy physical constraints; further, they do not explicitly model occluded regions. To circumvent these drawbacks, Tang \emph{et al.}~\cite{tang2017state} propose to refine the results of model-free method by inputting it to a physical simulator. Schulman \emph{et al.} ~\cite{schulman2013tracking} designed a modified expectation-maximization (EM)  algorithm and perform inference through calls to a physics simulator. These approaches are applied to track rope, sponge, and folded cloth.  In contrast, we are able to track the configuration of cloth in highly crumpled configurations, which has not been achieved in prior work.  We also demonstrate how model-based tracking can be used for self-supervised fine-tuning of a mesh reconstruction model.


\textbf{Closing the Gap Between Sim and Real}. 
Simulation has shown considerable promise for generating large amounts of labelled data at low cost, especially for domains where groundtruth supervision is difficult to obtain, such as optical flow and scene flow estimation~\cite{dosovitskiy2015flownet, ilg2017flownet, teed2020raft} or 3D reconstruction~\cite{park2019deepsdf, chi2021garmentnets}. However, models trained in simulation do not always readily transfer to the real-world due to the sim2real distribution shift.  Prior works~\cite{chi2019occlusion, huang2022mesh} shows that when applied to real world data, the performance of cloth reconstruction model drops significantly.
One strategy to bridge this gap (``sim2real") is to randomize the simulation parameters to create a diverse set of training data~\cite{sadeghi2016cad2rl, tobin2017domain, james2017transferring, akkaya2019solving}, with an underlying assumption that the randomized simulation data will cover the distribution of real data.
However, for cloth reconstruction, the source of distribution shift remains unclear. In other words, how to randomize cloth simulation properly might be a more difficult problem.

In this paper, we seek to resolve the distribution shift by fine-tuning a pre-trained simulation-trained model with real world data. The main challenges for fine-tuning a mesh reconstruction model in the real-world is the absence of ground-truth data (i.e., the full mesh). While there exists several real world datasets~\cite{bhatnagar2019multi, zhu2020deep} for on-body cloth reconstruction, directly obtaining the ground-truth full mesh of crumpled clothes in the real-world is very challenging due to self-occlusion,  i.e., the occluded portion of the clothes is not observable. To tackle this issue, we propose to generate pseudo labels using a action-conditioned tracking technique. Although the proposed tracking method relies on that the initial configuration of the cloth to be flattened, our learned cloth reconstruction model can be  applied to any cloth configurations.

\section{Method}
The goal of this project is to track the mesh of a cloth; we then use the tracked mesh to train a model to reconstruct the mesh of a (possibly crumpled) cloth  from a depth image observation. Past work in this area has trained a mesh reconstruction model in simulation and transferred the trained model to the real world~\cite{jiang2020bcnet,danvevrek2017deepgarment,saito2021scanimate,su2022deepcloth,chi2021garmentnets,li2014real,huang2022mesh,patel2020tailornet}.  However, such methods can suffer from a performance drop in the real-world due to the sim2real gap between simulated cloths and real cloths. To circumvent the issue, we design a self-supervised learning method for finetuning a mesh reconstruction model with unlabeled real data.

The high-level idea of our method is as follows: suppose that we know the full configuration of the initial mesh and a dynamics model of the cloth.  If we take an action on the cloth, then we can use the dynamics model to estimate the configuration of the cloth at the next timestep. However, since the dynamics model might not be perfect, we refine the prediction by aligning the predicted mesh with the observation through an optimization procedure. We view this procedure as similar to the motion update and measurement update of Bayesian filtering. 

Given an initial mesh reconstruction model trained in sim, the whole system can be divided into 3 stages: 
\begin{enumerate}
    \item Collect real-world trajectories. 
    
    \item 
    Track the motion of the cloth with our action-conditioned model-based tracking method.  
    \item Use the tracking output to generate pseudo-ground-truth (pseudo-gt) meshes and finetune the mesh reconstruction model. 
\end{enumerate}
We describe our method in more detail below.



\subsection{Data Collection in the Real-world}
\label{sec:method:data}
\begin{wrapfigure}{r}{0.23\textwidth}
    \vspace{-5mm}
    \centering

    \includegraphics[width=\linewidth]{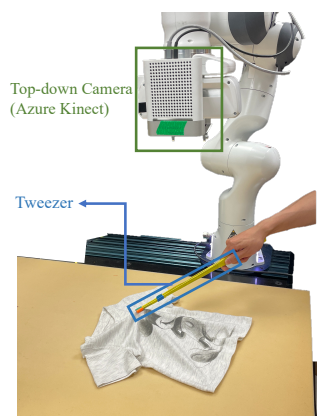}
    \caption{A human collector uses a tweezer to conduct pick-and-place actions. RGB-D videos are captured by a top-down camera.}
    \vspace{-3mm}
    \label{fig:data_col}
\end{wrapfigure}
 \begin{figure*}[h]
    \centering
    \includegraphics[width=1\textwidth]{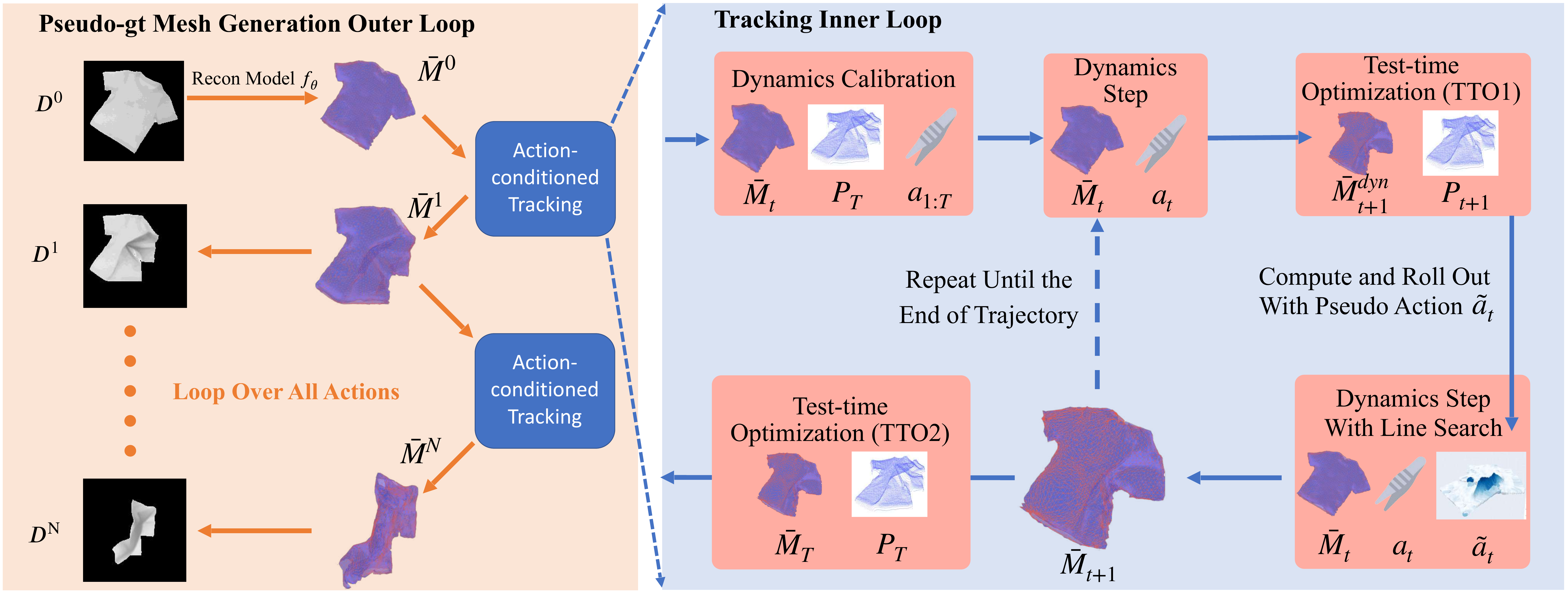}
    \caption{\textbf{Left}: The workflow for generating pseudo-ground-truth (pseudo-gt) meshes for one trajectory. We first reconstruct the initial mesh by using the pretrained mesh reconstruction model $f_{\theta}$. Then, we estimate the deformation caused by each pick-and-place action through action-conditioned tracking. By tracking all transitions sequentially, we obtain the pseudo mesh for crumpled cloths. \textbf{Right}: Before the tracking starts, we first run a parameter search to calibrate the dynamics model. Then we iterate over all low-level actions by: 1) rolling out the dynamics model with the picker action for one step; 2) running test-time optimization (TTO1) to align the model prediction result with observation, which produces a per-vertex "pseudo action"; 3) running the simulation again with a line search. After all the low-level actions within a pick-and-place action are executed, we run another optimization step (TTO2) to account for the tracking errors.}
    \label{fig:system}
    \vspace{-6mm}
\end{figure*}
In this section, we explain how we instrument and collect real-world trajectories.
In order to finetune the mesh reconstruction model, we need to collect real-world data of cloths in random configurations. We use a top-down camera placed above the workspace to capture all the observations. We initialize the state of the cloth into some configuration in which our mesh reconstruction model works reasonably accurately (such as a flattened state); we then perform a sequence of  pick-and-place actions. Then we reset the cloth into a new configuration in which our mesh reconstruction model works reasonably accurately  (e.g.,  another flattened state) and repeat.
When executing each action, we record a full RGB-D video including the intermediate states. The point cloud of the clothes is computed through color segmentation. The actions are conducted with a tweezer, which helps reduce the amount of occlusions compared to a robot gripper or human hand (Fig.~\ref{fig:data_col}). 

For each pick and place action, since we record the intermediate states,  
we obtain a  sequence of point clouds 
$P_{1:T}$
and low-level picker actions 
$a_{1:T}$.
Note that the entire sequence $a_{1:T}$ corresponds to intermediate actions within a single pick and place action.  We record a separate action and point cloud sequence for each pick and place action.  In our experiments, we 
apply 3 pick-and-place actions ($N=3$) in each trajectory, which takes around one minute (per trajectory) for an experienced human collector; after each 3 pick-and-place actions, we reset the cloth to a flattened state.

\subsection{Pseudo-gt Mesh Estimation by Model-based Cloth Tracking}
\label{sec:method:tracking}
Given an initial depth image $D^0$, a pretrained mesh reconstruction model, an (imperfect) dynamics model, as well as the action and point cloud sequences recorded in the previous section, our next goal is to estimate the full mesh for every state recorded in the dataset. 
We assume that the initial depth image $D^0$ is recorded from the cloth in a flattened state in which the pre-trained mesh reconstruction model is reasonably accurate.  Let us denote the estimated reconstruction of this initial mesh 
as $\Bar{M}^0$, where a mesh is defined as $M=(V, E)$ with vertices $V$ and edges $E \subseteq V \times V$. 


Given the initial estimated mesh $\Bar{M}_0$, a sequence of point clouds $P_{1:T}$, and a sequence of actions, $a_{1:T}$, our objective is to estimate a sequence of meshes corresponding to each timestep $\Bar{M}_{1:T}$. 
To obtain an accurate estimate of the motion of the cloth, we developed an action-conditioned model-based cloth tracking method that is robust to occlusions. 
We first simulate each action with the imperfect dynamics model to obtain an initialization of the motion.  We then run an optimization to match the visible mesh with the observed point cloud.  Finally, we use the dynamics model again to obtain the final prediction (see Fig.~\ref{fig:system}).

\subsubsection{Initialize the Motion with a Dynamics Model}
\label{sec:method:tracking:dyn}
Our method falls under ``model-based tracking"~\cite{schulman2013tracking,tang2017state,wang2021tracking}.
Compared to model-free tracking, one of the most appealing properties of model-based tracking is that it models the whole object, including the occluded part. This is significant when we track double-layer cloths, because part of the cloths might be occluded throughout the entire trajectory. In this case, model-free tracking is only able to estimate the motion of the visible part while leaving occluded portion of the mesh unchanged. On the other hand, model-based tracking can use a physics prior of cloth to estimate the motion of the occluded regions and estimate the configuration of the full mesh that satisfies physical motion constraints. 

At each timestep $t$, we directly modify the position of the picked particle according to the recorded picker action, $a_t$.  We then run the dynamics model $dyn$ for one step to propagate the effect to the whole cloth, holding fixed the position of the picked particle. Suppose $x_t\in \mathbb{R}^{|V|\times3}$ are the positions of all vertices; given the vertices $x_{t}$ and action $a_{t}$, the dynamics model will predict the next state $x_{t+1}^{dyn}=dyn(x_{t}, a_{t})$.
We define $\Delta x^{dyn}_{t+1}= x^{dyn}_{t+1}-x_{t}$ to denote the motion of all vertices, which will be used as an initialization for the test-time optimization, as described in Sec.~\ref{sec:method:tracking:measure}.

In order to make the dynamics model as realistic as possible, we calibrate the dynamics model by searching for the optimal physical parameters (such as friction and stiffness). To do so, we first simulate the entire action sequence $a_{1:T}$ to obtain the final predicted mesh $\hat{M}_T$.  We then use a z-buffer to compute the visible portion of the predicted mesh $\hat{M}_T^{vis}$.  Next, we run a grid search over the dynamics model parameters to minimize the Chamfer distance between the visible portion of the simulated mesh $\hat{M}_T^{vis}$ and the point cloud at the final step $P_T$. Then we use the optimized parameters to obtain $x_{t+1}^{dyn}$ as explained above. Dynamics calibration is carried out in an online fashion: we calibrate the dynamics model separately for each pick-and-place action. We show the necessity of online calibration in Appendix Sec. 2.1 (see our \href{https://sites.google.com/view/ss-mesh-recon/home}{website}).
For our dynamics model, we use a position-based cloth dynamics model implemented in the Nvidia FleX simulator~\cite{corl2020softgym, ha2022flingbot}.


\subsubsection{Augment Imperfect Dynamics Model by Aligning with Measurement}
\label{sec:method:tracking:measure}
Due to the complexity of real-world dynamics and the challenges of system calibration, our dynamics model will have errors. Even with accurate estimation of the initial state, it will deviate from the real-world rollout with errors accumulating over time.
To tackle this challenge, we draw inspiration from the measurement update step of Bayesian Tracking~\cite{arulampalam2002tutorial}:
we eliminate the compounding errors due to the inaccurate dynamics model by running a test-time optimization (TTO1 in Fig.~\ref{fig:system}) and thereby reduce the discrepancy between the dynamics prediction and the measurement (the observed pointcloud).

From the dynamics model (Sec.~\ref{sec:method:tracking:dyn}), we obtain an initial estimate of the state of cloth $x_{t+1}^{dyn}$. The goal of the test-time optimization step is to compute a correction term $\Delta x_{t+1}^{corr}$ that adjusts the predicted mesh to better match the observed point cloud.  We optimize a 3-D translation for each vertex $\Delta x_{t+1}^{corr}$ so that $x_{t+1} = x_{t+1}^{dyn} + \Delta x_{t+1}^{corr}$ is aligned with the observation. The specific optimization objectives are:

\textbf{Chamfer Loss}. The first objective we have is the one-way Chamfer distance~\cite{huang2022mesh, sundaresan2022diffcloud} between the next point cloud $P_{t+1}$ and the visible portion of the mesh $x_{t+1}^{vis}$, given by $\mathcal{L}_{Chamf} = \mathcal{D}_{chamf}(P_{t+1}, x_{t+1}^{vis})$. We use the one-way Chamfer distance because the observed point cloud is incomplete due to occlusions induced by the tweezer. For each point on the observed point cloud, we find the nearest neighbor within the visible set of mesh vertices. 

\textbf{Rigidity Loss}. As-Rigid-As-Possible (ARAP)~\cite{sorkine2007rigid,su2015render,park2021nerfies,slavcheva2017killingfusion} is a common assumption for modeling cloth-like shapes. In \emph{TTO1}, the correction term $\Delta x_{t+1}^{corr}$ can be viewed as motion that transforms the mesh to match the partial point cloud at the next timestep. Based on ARAP, we assume the cloth nodes in neighboring regions move as rigidly as possible. Intuitively, this loss helps improve the consistency of motions of adjacent particles. Since we only model the motion by a translation ($\Delta x_{t+1}^{corr}$), we obtain a simplified rigidity loss:

\begin{equation}
    \begin{aligned}
    \mathcal{L}_{Rig} = \frac{1}{|E_t|} \sum_{i,j \in E_t} \left\| \Delta x^{corr}_{t+1, i} - \Delta x^{corr}_{t+1, j} \right\|_2^2
    \end{aligned}
\end{equation}

At each step, we optimize $\Delta x_{t+1}^{corr}$ with respect to Eq.~\ref{eq:tto_obj} for 200 iterations with the Adam optimizer~\cite{kingma2014adam}. In our experiment, we set $\alpha=1$ and $\beta=10$.
\begin{equation}
    \begin{aligned}
    \argmin_{\Delta x_{t+1}^{corr}} \alpha \mathcal{L}_{Chamf} + \beta \mathcal{L}_{Rig}
    \end{aligned}
    \label{eq:tto_obj}
\end{equation}

\subsubsection{Rollout Augmented Dynamics with Line Search}
\label{sec:method:tracking:ls}
In the previous section, we described how to compute a correction term for the predicted mesh based on the observed measurement. However, 
the optimized mesh is not guaranteed to satisfy all physical constraints. Similar to~\cite{tang2017state, tang2018track}, we use the dynamics model again to verify the physical plausibility of the mesh.
We rollout the dynamics model again from the original state $x_{t}$; this time, we use both the picker action $a_{t}$ as well as a ``pseudo-action"  $\tilde{a}_t =   \Delta x_{t+1}^{dyn} + \Delta x_t^{corr}$. The picker action is executed as described in Sec.~\ref{sec:method:tracking:dyn}.
This time, the pseudo-action $\tilde{a}_t$ is applied to all visible particles (not just the picked particle).
We then use the dynamics model to adjust the positions of the occluded particles.


Although the pseudo action $\tilde{a}_t$ helps align the rollout with observation, it may potentially create physically infeasible configurations. Therefore, we run a line search on the correction component $\Delta x_t^{corr}$.
If the simulation explodes, i.e., the velocities of cloth particles exceed a pre-defined threshold, we multiply $\Delta x_t^{corr}$ with a decaying factor $\gamma$ (we set $\gamma = 0.7$). If the simulation fails for 10 times, we set $\tilde{a}_t = 0$, which means the pseudo action is not used. 


As shown in Alg.1 line~\ref{alg1:line4}-\ref{alg1:line8}, we iterate over all action segments to track a single pick-and-place action. In our experience, even after the pseudo-action, the tracking result (estimated mesh) may still deviate from the observation due to compounding errors. To ensure that the estimated mesh is well aligned with the observation, we run another test-time optimization (TTO2) with an identical set of losses as before. 
Finally, we use the optimized mesh as the initial state for the next pick-and-place action 
and iterate until all pick-and-place actions have been tracked. We ablate TTO2 in Appendix Sec. 2.2 (see the \href{https://sites.google.com/view/ss-mesh-recon/home}{website}) and show that it also improves the robustness to the errors of the dynamics model.

\begin{center}

\begin{algorithm}
\SetAlgoLined
\DontPrintSemicolon 
 \SetKwInOut{Input}{Input}
 \SetKwInOut{Output}{Output}
 \Input{A sequence of depth image sequences $\{D_{1:T}\}_{i=0}^N$, picker actions  $\{a_{1:T}\}_{i=0}^N$, point cloud sequence $\{P_{1:T}\}_{i=0}^N$, a pretrained mesh reconstruction model $f_\theta$, and a dynamics model $dyn$. } 
 \Output{A pseudo-labeled dataset that includes paired observations and pseudo-gt mesh: B=$\{(D_i, \bar{M^i})\}_{i=0}^N$}
Reconstruct initial mesh $\bar{M}^0_0$ by pre-trained model $f_\theta$ \; \label{alg1}
Initialize the pseudo-labeled dataset B with $(D^0, \bar{M}^0_0)$\;
 \For{$i \gets 0$ \textbf{to} $N$} 
 {

    \For{$t \gets 1$ \textbf{to} T}{         \label{alg1:line4}
        $M_{t}^{dyn,i} \gets dyn(\bar{M}^i_{t-1}, a^i_{t})$  \;  \label{alg1:line6}
        $\tilde{a}_t^i \gets $ GetPseudoActionByTTO($M_{t}^{dyn,i}, P^i_t$)  (Sec.~\ref{sec:method:tracking:measure}) \;  \label{alg1:line7}
        $\bar{M}_{t}^i \gets dyn(\bar{M}_{t-1}^i, a_t^i, \tilde{a}_t^i)$ \textbackslash \textbackslash with line search
    }                                       \label{alg1:line8}
    Optimize $\bar{M}_T^i$ with test-time optimization \; \label{alg1:line9}
    Add depth image $D^{i}_T$ and pseudo mesh $\bar{M}^{i}_T$ to the pseudo-labeled dataset B\;
    Use the final mesh in the current iteration as the initialization of next iteration: $\bar{M}^{i+1}_0 \gets \bar{M}_T^i$ \;
 }
 
 \Return{Pseudo-labeled dataset B}
 \caption{Pseudo-gt mesh generation}
\end{algorithm}
\end{center}
\vspace{-8mm}
\subsection{Model finetuning}
Using the above tracking model, we obtain a pseudo-labeled dataset, with an estimated mesh for each observed point cloud or depth image.  Our dataset consists of the final estimated mesh at the end of each pick and place action $\bar{M}_T$ and the corresponding point cloud $P_T$.
After curating this pseudo-ground-truth dataset, we  use it to finetune the mesh reconstruction model. 
In our experiments, we finetune the mesh-reconstruction model in MEDOR~\cite{huang2022mesh}, which is built off GarmentNets~\cite{chi2021garmentnets}. Given a depth image, GarmentNets~\cite{chi2021garmentnets} and MEDOR~\cite{huang2022mesh} first predict the canonical coordinates of each pixel.  They then complete the shape in canonical space and finally transform the completed shape back to observation space. It should be noted that our method not only provides the pseudo ground-truth mesh for the observation space, but our method also can compute a pseudo-ground-truth mesh in the canonical space. This is because GarmentNets~\cite{chi2021garmentnets} and MEDOR~\cite{huang2022mesh} simultaneously reconstruct both meshes in the initial configuration; tracking the meshes helps preserve the mapping between canonical space and observation space. Thus, we are able to fine-tune both the model that maps from observation space to canonical space as well as the model that maps from canonical space back to the observation space (i.e. both parts of the mesh reconstruction model).
In terms of the GarmentNets~\cite{chi2019occlusion} components, we train the canonicalization model, as well as the shape completion and warp field prediction models. For details, please refer to GarmentNets~\cite{chi2019occlusion}.




\section{Experiments}
Through the experiments, we seek to the answer the following questions: 
\begin{enumerate}
    \item Can our method generate approximately correct pseudo-gt meshes for mesh reconstruction and dynamics learning?
    \item Can our model adapt quickly after being finetuned on the pseudo-gt meshes?
\end{enumerate}

\subsection{Evaluation on the Quality of Pseudo Labels}

\textbf{Setup}. We collect 50 trajectories in the real world, each of which contains 3 pick-and-place actions. Including the initial state, there are 200 pseudo labels in total. 

\textbf{Baselines}. We compare our method to 4 baselines:
\begin{itemize}
    \item \textbf{No Pseudo Action}. The goal of pseudo action $\Tilde{a}_t$ is to ``patch'' the inaccuracies of the dynamics model. We verify its effectiveness by removing it from the method and only using the recorded picker action $a_{1:T}$. 
    \item \textbf{No Action Conditioning}. In this baseline, we assume the picker action information is not known, which has two implications. 1) When computing the pseudo action, since we don't know the picker action, we cannot roll out the simulator to initialize TTO; 2) When we simulate the pseudo action with line search, we only apply the pseudo action alone (not the picker action).
    \item \textbf{No Dyn Init}. In this baseline, we directly run TTO on the current mesh $M_t$ instead of the simulated next mesh $M_{t+1}^{dyn}$. This is to verify whether using the dynamics model to bootstrap the optimization is critical to the performance.
    \item \textbf{No Test-time Finetuning (TTO2)}. Although we conduct TTO in between the dynamics rollout, the rollout may still drift due to imperfect dynamics. In this baseline, we remove the optimization step at the end of the tracking procedure (TTO2) to see whether this component is necessary.
    
\end{itemize}
\vspace{-2.mm}
\textbf{Metrics}. Evaluating the quality of the pseudo label is challenging, due to the absence of the ground-truth mesh. To evaluate the quality of the pseudo label, we use the bidirectional Chamfer distance between the visible surface of the pseudo mesh and the observed point cloud. Our assumption is that if the tracking is accurate, then the visible surface of the pseudo mesh should match the observed point cloud, which is the visible surface of ground-truth mesh. We compute the metric with the mesh before Test-time Optimization 2. This is because,  even for a completely erroneous prediction, the shape can match the observation after TTO2 and achieve a low cost. Therefore, comparing the loss after TTO2 is not very indicative of accurate tracking.
\begin{wraptable}{r}{5.2cm}
\vspace{-0.4cm}
\scriptsize
    \centering
    \begin{tabular}{c|c}
        \toprule
        \multirow{1}{*}{Method} & \multicolumn{1}{c}{Chamfer PC $\downarrow$}  \\ 
            & ($1\times 10^{-4}$ m)\\
            \midrule
            No Pseudo Act (No TTO1)  &   $1.62\pm 0.98  $ \\ 
            No Dyn Init & $1.40 \pm 1.21$ \\
            No Act Cond & $3.72 \pm 2.32$ \\
            No TTO2 & $2.19\pm 1.69$\\
        \hline
        MEDOR & $3.0\pm 2.1$\\
        \textbf{Ours (full method)} & $\mathbf{1.13\pm 1.24}$\\
        \hline
    \end{tabular}
\caption{Quantitative results of different variants of our method.}
\label{table:label_quality}
\vspace{-.4cm}
\end{wraptable}

\begin{figure*}[t]
    \centering
    \includegraphics[width=\textwidth]{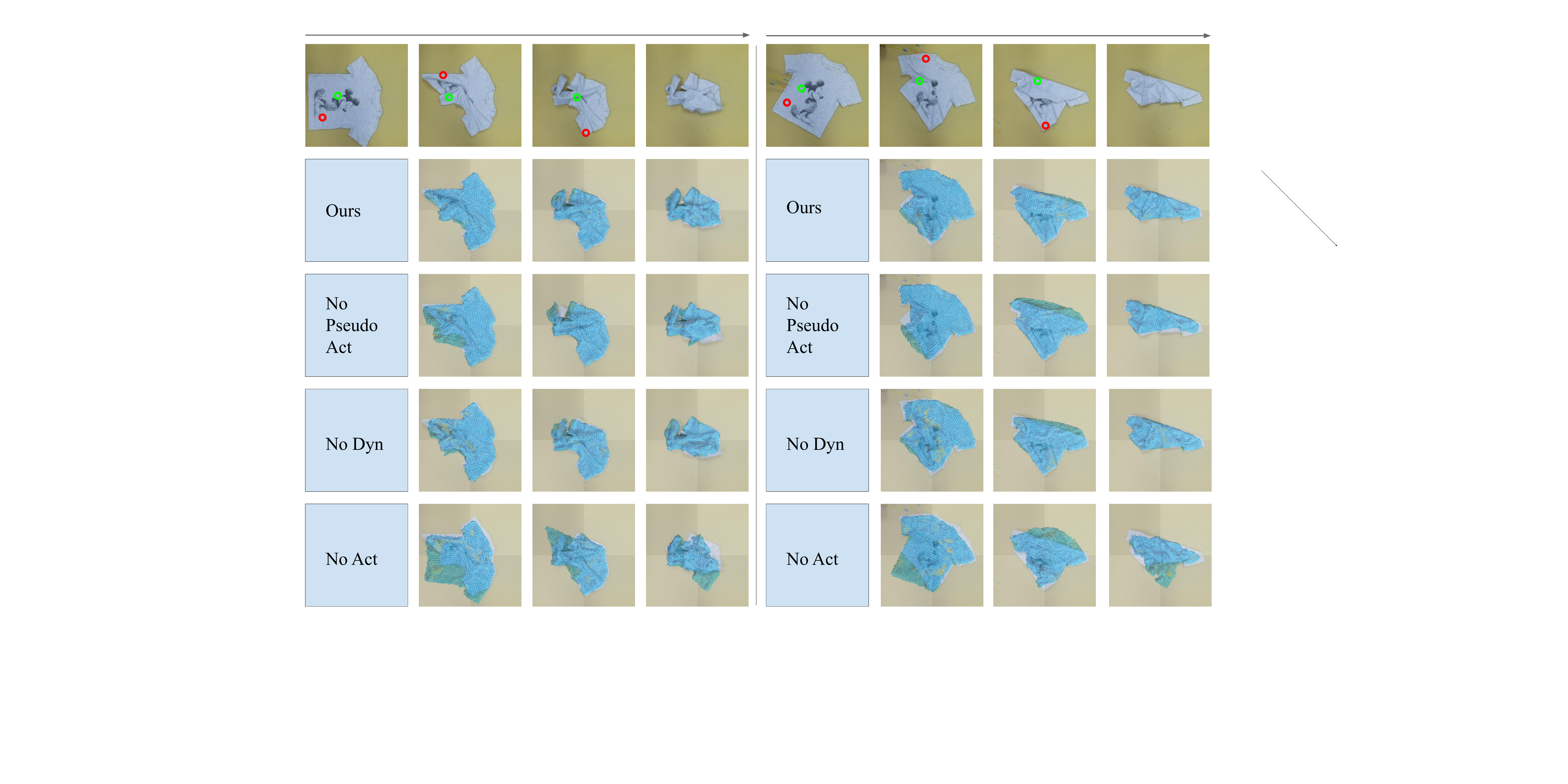}
    \caption{Qualitative results of pseudo-gt mesh generation. First row is the real-world rollout, with pick points and place points denoted by red and green circles respectively.}
    \label{fig:example_traj}
    \vspace{-6mm}
\end{figure*}

\textbf{Results}. In the Fig.~\ref{fig:example_traj}, we show the side-by-side 
comparison of the tracking results of the different methods. Videos and 3D visualizations of the pseudo mesh can be found on our
\href{https://sites.google.com/view/ss-mesh-recon/home}{website}. Our full method (second row) is able to track the clothes even under complicated configurations, i.e., multiple folds (right figure). In contrast, the other methods all produce pseudo meshes that don't align with the observations. 

In Table.~\ref{table:label_quality}, we show the quantitative results of the baselines and our method. Means and standard deviations of the generated pseudo meshes are computed across the whole dataset. Comparing our method with \emph{No Pseudo Act}, we see that dynamics errors can be significantly reduced by aligning with measurements during the rollout. By comparing \emph{No Dyn Init} with our method, we see the importance of using dynamics prior as the initialization for the optimization problem. Since tracking is a correspondence problem, our conjecture is that initializing with a dynamics rollout makes it easier to find correspondences. Looking at \emph{No Act Cond}, we see the benefits for tracking deformable objects conditioned on the action, and a pure optimization method failed in this case.  Comparing \emph{No TTO2} with our method, we observe that TTO2 helps reduce the compounding error; thus removing it hurts the performance.

\subsection{Model Performance After Finetuning}



In this section, we investigate whether the pseudo-gt mesh generated from our proposed workflow is beneficial for the self-supervised learning of a mesh reconstruction model.  We finetune MEDOR~\cite{huang2022mesh} which is purely trained in simulation, and show its performance before and after finetuning.

\subsubsection{Mesh reconstruction}
\begin{table}
    \centering
    \begin{tabular}{@{}c|c|c|c@{}}
        \toprule
        Method & Chamfer PC $\downarrow$ & Chamfer Mesh $\downarrow$ & Flattening $\uparrow$\\
        \midrule
        MEDOR~\cite{huang2022mesh} w/o ft & $3.0\pm2.1$ & $2.2\pm1.7$ & 0.23\\
        MEDOR~\cite{huang2022mesh} w/ ft & $\mathbf{1.6\pm 1.4}$ & $\mathbf{1.2\pm0.8}$ & $\mathbf{0.3}$\\
        \hline
    \end{tabular}
\caption{Performance before and after finetuning.}
\label{table:ft_perf}
\vspace{-6mm}
\end{table}
In Table~\ref{table:ft_perf}, we show 2 metrics. The \emph{Chamfer PC} is the bidirectional Chamfer distance between the visible predicted mesh and the partial point cloud, which is the same metric as in Table.~\ref{table:label_quality}. \emph{Chamfer PC} indicates how well the prediction aligns with the observation. \emph{Chamfer Mesh} is the chamfer distance between the full predicted mesh and the pseudo mesh, which indicates how well the model learns from the pseudo labels. As we can see from  Table~\ref{table:ft_perf}, both metrics are significantly improved after finetuning ($46\%$ for \emph{Chamfer PC} and $45\%$ for Chamfer Mesh).

\subsubsection{Robot Cloth Flattening}
In order to demonstrate the effectiveness of our method in robotic manipulation, we also deployed the finetuned model for a physical robot experiment: robot cloth flattening. The goal of this task is to maximize the coverage of a T-shirt by using a 
7-DoF Franka robot and pick-and-place action. We use normalized improvement as the metric: 0 means no improvement and 1 means the T-shirt is completely flattened. Following MEDOR~\cite{huang2022mesh}, we integrate the fine-tuned mesh reconstruction model with a learned mesh-based dynamics model for planning. The details of the task can be found in the appendix.

We test the model with and without finetuning for 6 trajectories separately, and calculate the average normalized improvement. Each trajectory contains 10 pick-and-place actions.  We observe a performance gain of 30.4\% after finetuning with the pseudo-labeled dataset (Table~\ref{table:ft_perf}). This shows that the quality of pseudo mesh is sufficiently accurate for improving the downstream manipulation task.


\vspace{-1mm}
\section{Conclusions}
We proposed a self-supervised mesh reconstruction method in the real world, via  action-conditioned cloth tracking. We show that by leveraging a dynamics model and optimization, we can accurately track  cloth and compute pseudo-labels of the reconstructed mesh for crumpled cloths. By finetuning a simulation-trained mesh reconstruction model on the real-world pseudo labels, we can partially close the sim2real gap and improve the performance of cloth reconstruction and manipulation in the real world.



\section*{ACKNOWLEDGMENT}
This work was supported by LG Electronics, National Science Foundation (NSF) CAREER Award (IIS-2046491) and NSF Smart and Autonomous Systems Program (IIS-1849154).


\bibliographystyle{IEEEtran}
\bibliography{main}

\newpage
\begin{center}
    
    {\Large Supplementary Material}  
\end{center}

\section{Real-world Cloth Flattening}
In order to further demonstrate the potential of our self-supervised mesh reconstruction method for robotic application, we deploy it in real world for a cloth flattening task.

\subsection{Experiment Setup}
The objective of the experiment is to flatten a crumpled Tshirt by using a 7-DoF Franka robot and pick-and-place action. The evaluation metric is the normalized improvements of coverage (0 if no changes, 1 if maximum coverage is reached). Since our goal is to evaluate whether the pseudo label sufficiently accurate to improve the performance of manipulation task, we use the same Tshirt for flattening as we collect the pseudo label dataset.

\subsection{Model-based Cloth Manipulation System}
After finetuning the mesh reconstruction model with pseudo label dataset, we integrate it with a learned graph dynamics model for planning. At each step, we first reconstruct the cloth with mesh reconstruction model. Then we sample 100 random pick-and-place actions and roll out with the dynamics model. We use cloth coverage  as the reward function and execute the action the results in highest coverage.


\section{Ablation}
\subsection{Online vs offline dynamics calibration}
\begin{table}{f}
    \vspace{-0.4cm}
    \centering
    \caption{Ablation on the necessity of online simulation calibration.}
    \begin{tabular}{c|c}
        \toprule
        \multirow{1}{*}{Method} & \multicolumn{1}{c}{Chamfer PC}  \\ 
            & ($1\times 10^{-4}$)\\
            \midrule
        Online (Ours) & $1.13\pm1.24$ \\ 
        Offline & $1.90\pm 1.53$ \\
    \end{tabular}
    
    \label{table:abl_online}
\end{table}

Due to the simplification of dynamics model and complexity of real world environment, it's difficult to find a single set of simulation parameters that work well for different configurations. 
In this section, we investigate the necessity of online simulation calibration. 

\textbf{Online dynamics calibration}: identify the dynamics parameters for each pick-and-place actions separately, in an online fashion. We adopt online dynamics calibration in our main method.

\textbf{Offline dynamics calibration}: identify the modes of the dynamics parameters on an offline dataset and transfer them to individual trajectories. In our experiment, we find the modes of dynamics parameters on the entire dataset

\subsection{Ablation on Test-time Optimization 2 (TTO2)}
\begin{figure*}[t]
    \centering
    \includegraphics[width=\textwidth]{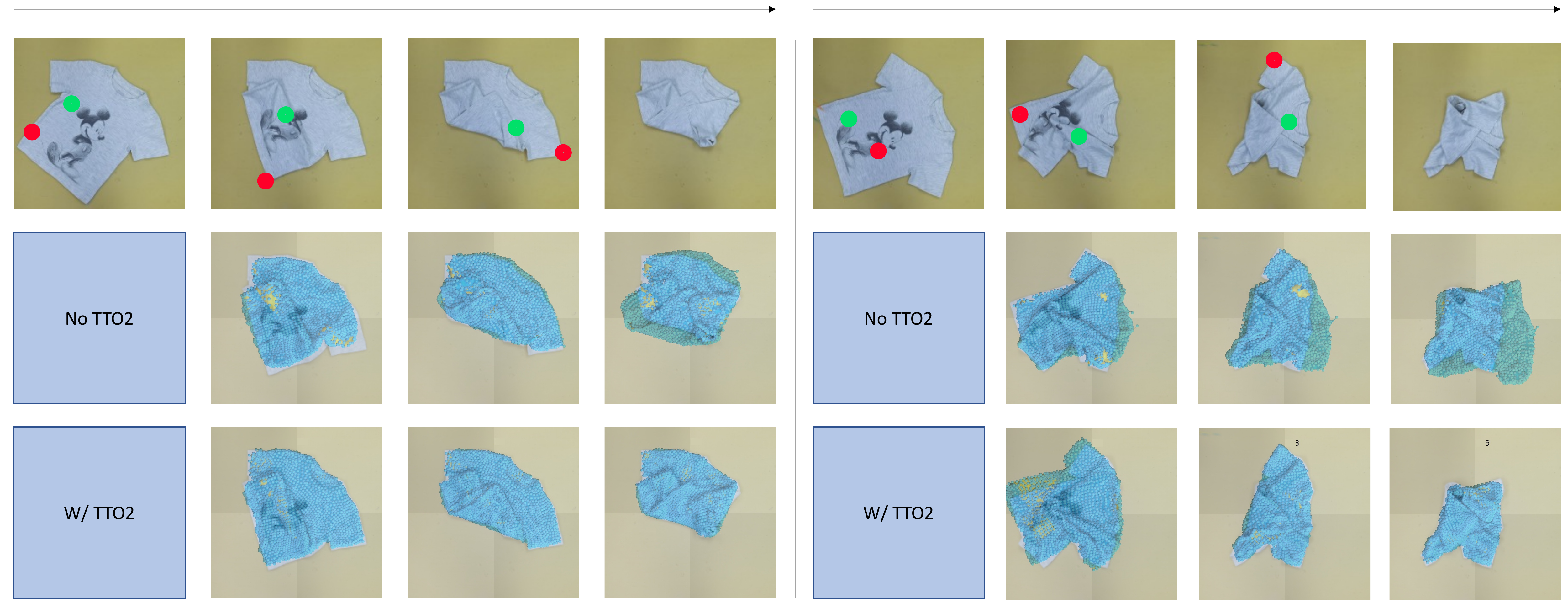}
    \caption{Qualitative results for ablation on TTO2. After removing TTO2 (second row), the errors compounded over the pick-and-place actions. The final mesh (4th column) notably deviates from the observation.}
    \label{fig:tto2_qual}
\end{figure*}

In Fig.~\ref{fig:tto2_qual}, we show a qualitative comparison between with and without TTO2: TTO2 alleviates the compounding error over several pick-and-place actions. Additionally, we also find that TTO2 minimizes the need for a good model. We conduct an ablation to verify this assumption. During simulation calibration, instead of choosing the best simulation parameters, we intentionally choose parameters that result in a worse dynamics model. As shown in the table below, we found that TTO2 improves the robustness of our method towards model quality. The column ``Top 50\%`` or ``Top 90\%`` refers to the ranking of the dynamics parameters that we have sampled. As shown, with TTO2, there is only a drop of 19.5\% when using the incorrect dynamics parameters (top 90\%) compared to using the best parameters; without TTO2, there is a much larger drop of 30.1\% when using the incorrect dynamics parameters.

\begin{table*}[h]
    \centering    \scriptsize
    \begin{tabular}{c|c|c|c|c}

    \toprule
     \diagbox{Method}{\ Model Quality} & Best & Top 50\% & Top90\% \\ \hline
    w/o TTO2 & $2.19\pm 1.69$ & $2.67\pm2.04$ &$2.85\pm 2.32$ \\
 w/ TTO2 & $1.13\pm 1.24 $    & $1.18\pm 1.20$  & $1.35\pm 1.29 $  \\

    \bottomrule
    \end{tabular}
    \caption{TTO2 improves the robustness to model error.
}
    \label{tab:TTO_abl}
\end{table*}

\subsection{Qualitative results of fine-tuned model}
\begin{figure*}[t]
    \centering
    \includegraphics[width=\textwidth]{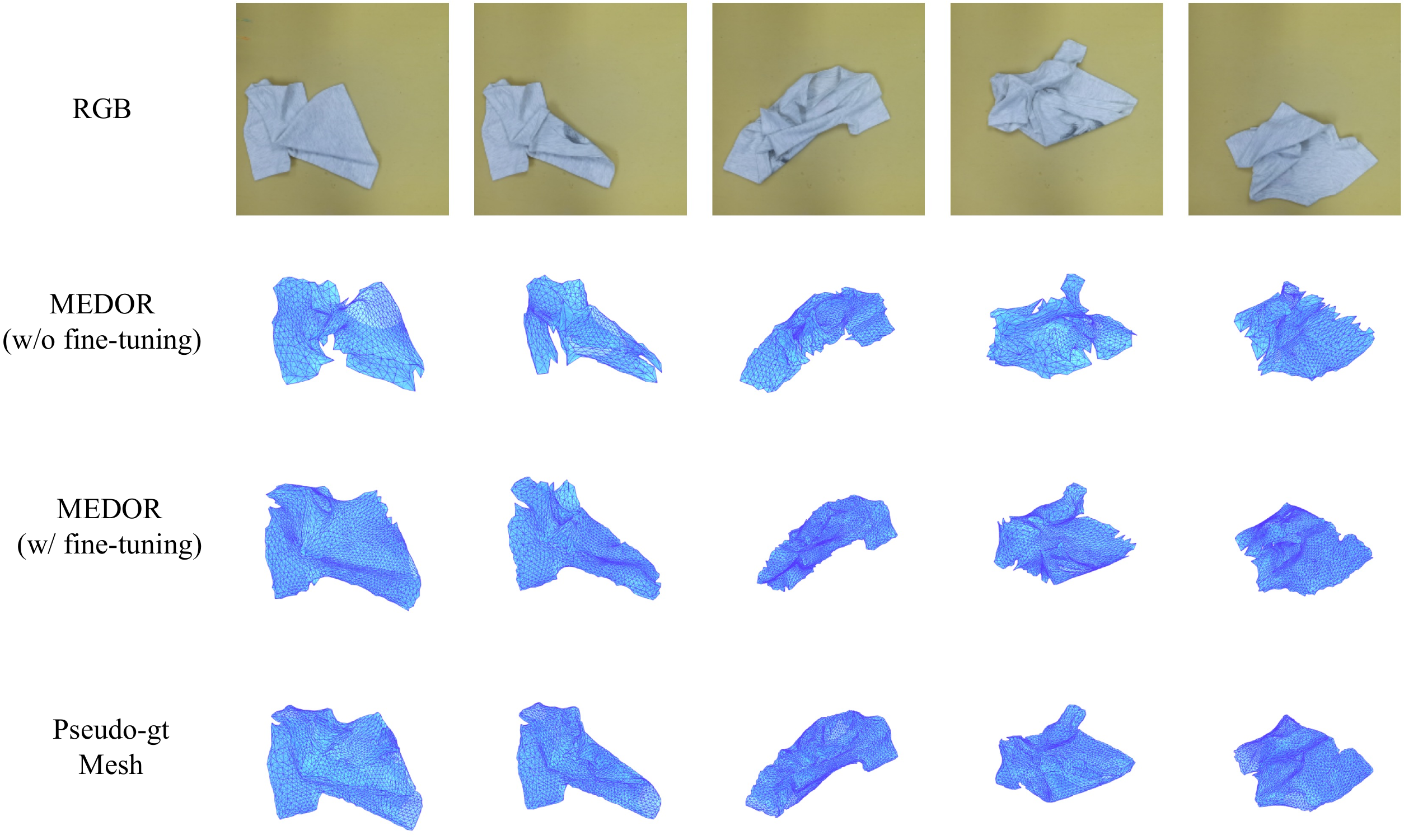}
    \caption{Qualitative results for ablation on TTO2. After removing TTO2 (second row), the errors compounded over the pick-and-place actions. The final mesh (4th column) notably deviates from the observation.}
    \label{fig:model_ft_qual}
\end{figure*}

In Fig.~\ref{fig:model_ft_qual}, we visualize the results of state of the art cloth reconstruction model, MEDOR~\cite{huang2022mesh} (2nd row), and MEDOR after being finetuned (3rd row) by the pseudo-gt mesh (4th row). It shows that our self-supervised approach can reliably generate pseudo-gt mesh from partial observation (depth image). This pseudo-gt mesh can be used for finetuning cloth reconstruction model and improves its performance in real-world.

\subsection{Ablation on Collision}
To better motivate the adoption of rigidity loss, we conduct a collision test on the pseudo-gt mesh, generated with or without rigidity loss in TTO1 and TTO2. We define a ``collision`` as when the distance between two vertices is less than a predefined threshold (0.005). In Nvidia Flex, the distances between adjacent vertices are set to be the particle radius by default. Therefore, we use the particle radius (0.005) as the threshold. The average number of vertices for the pseudo-gt mesh is 3,906. Without rigidity loss, there are 33,765 pairs of collisions. After adding rigidity loss, the average number of collisions reduces to 4,099, which is approximately 9 times less frequent.

\section{Additional Details}
\subsection{Simulation Calibration}
\begin{table*}{h}
    \centering
    \begin{tabular}{c|c}
        
        Parameters & Range\\
        \midrule
        Stiffness & [0.2, 0.55, 0.9, 1.25, 1.6] \\ 
        Dynamic Friction Coefficient & [0.5, 1.4, 2.3, 3.2, 4.1, 5] \\
        Particle Friction Coefficient & [0.5, 1.4, 2.3, 3.2, 4.1, 5]\\
    \end{tabular}
    \caption{Types and range of physical parameters that we optimize during simulation calibration phase.}
    \label{table:sim_cal}
\end{table*}
Before we start to track to motion of cloth, we firstly calibrate the simulation by identifying the values of several critical physical parameters. Due to the simplified dynamics of simulation, one may not able to find a single set of parameters that allow the simulation to match real world in every possible transitions. Therefore, for each pick-and-place action, we search for the optimal system parameters that best simulate the current action.

We use Nvidia Flex as our simulator, and we find clothes stiffness and friction to be the most parameters.
During the simulation calibration, we directly roll out the dynamics model with actions $a_{1:T}$, without any bells and whistles. We run a grid search over all combinations of parameters (see Table.~\ref{table:sim_cal}). On a single Nvidia GTX 2080Ti, it takes around 70 seconds to run over the 125 combinations of parameters.

\subsection{Test-time Optimization}
Test-time Optimization (TTO) is an important component in our framework. It is applied twice in our action-conditioned tracking pipeline. TTO1 is applied iteratively inside the simulation loop of tracking process. The main goal of TTO1 is to augment the dynamics model by computing a pseudo action that aligns the simulated result with the measurement. Due to the inevitable gap between real world and simulation, it is possible that simulation cannot fully match the real world even with the help of pseudo action. For example, if the clothes in the simulation is thicker than the real world's, then the simulated mesh will always differ from the real mesh, otherwise the physics constraint will be violated. Therefore, after the inner simulation loop, we apply another test-time optimization, which we refer as \emph{TTO2}.

\subsection{Finetuning for MEDOR}
MEDOR~\cite{huang2022mesh, chi2021garmentnets} consists of 3 components, a canonicalization network that maps pixel from observation space to canonical space, a implicit shape completion network that predicts winding number field~\cite{jacobson2013robust}, and a warp field prediction network that predicts a per-vertex transformation from canonical pose to observation space. The model is finetuned in a two-stage process similar to training~\cite{huang2022mesh,chi2021garmentnets}.

In the first stage, we train the canonicalization network alone. It should be noted that at the beginning of the tracking procedure, we use a pretrained MEDOR model to reconstruct the flattened mesh. This can be seen as registrating the mesh to canonical space because we have the correspondence between observation space to canonical space. Then, by tracking the positions of vertices in the subsequent steps, we obtain the pseudo training label for the canonicalization network.

In the second stage, we train the shape completion network, and warp field prediction network with the reconstructed mesh in the canonical space and observation space separately. We use Adam~\cite{kingma2014adam} optimizer with cyclic learning rate~\cite{smith2017cyclical} between $1e^{-5}$ and $1e^{-6}$. The model is trained for 1000 epochs in the first stage and 2000 epochs in the second training stage.For model finetuning, we split the trajectories randomly into train and test set by a ratio of 9:1. Each trajectory contains 3 pick-and-place actions, which contains 3 crumpled cloth configurations.

\newpage

\end{document}